\documentclass[11pt]{article}

\usepackage[final]{acl}

\usepackage{times}
\usepackage{latexsym}
\usepackage{tcolorbox}

\usepackage[T1]{fontenc}

\usepackage[utf8]{inputenc}

\usepackage{microtype}

\usepackage{inconsolata}

\usepackage{graphicx}
\usepackage{enumitem}
\usepackage{amsmath}
\usepackage{multirow}
\usepackage[table,xcdraw]{xcolor}
\usepackage{booktabs}
\usepackage[normalem]{ulem}
\usepackage{listings}
\usepackage{fontawesome}
\useunder{\uline}{\ul}{}
\lstdefinestyle{PromptStyle}{
    basicstyle=\scriptsize\ttfamily\color{black!85},
    frame=lr,
    rulecolor=\color{black!60},
    backgroundcolor=\color{black!2},
    framesep=4pt,
    breaklines=true,
    breakatwhitespace=false,
    columns=fullflexible,
    numbers=none,
    aboveskip=6pt,
    belowskip=6pt,
    xleftmargin=2pt,
    xrightmargin=2pt,
    upquote=true
}

\title{From Empathy to Personalized Empathy: Adapting Empathetic Strategies to Individual Users}

\author{
 \textbf{Wuqiang Zheng\textsuperscript{1}},
 \textbf{Chengbing Wang\textsuperscript{1}\thanks{Corresponding author}},
 \textbf{Yilin Yang\textsuperscript{1}},
 \textbf{Junyi Cheng\textsuperscript{2}\footnotemark[1]},
\\
 \textbf{Jianfei Xiao\textsuperscript{1}},
 \textbf{Hu Sun\textsuperscript{2}},
 \textbf{Yi Xie\textsuperscript{2}},
 \textbf{Yangyang Li \textsuperscript{3}},
 \textbf{Wenjie Wang\textsuperscript{1}}
\\
\\
 \textsuperscript{1}University of Science and Technology of China,
\\
 \textsuperscript{2}Huawei Technologies,
 \textsuperscript{3}China Academy of Cyber
\\
 \small{
   \{qqqqqzheng, wcb0219\}@gmail.com
 }
}

\begin{document}
\maketitle

\begin{abstract}
As Large Language Models (LLMs) are increasingly deployed in long-term interactions with users, empathy has become an increasingly important capability.
However, existing research overlooks the influence of users' personality traits on empathetic strategies during long-term interactions. 
To address this gap, we introduce the task of personalized empathy, which focuses on adapting empathetic strategies according to users' personalized characteristics derived from history.
To study and enhance this capability, we construct PersonaEmp, a personalized empathy dataset built from long-term user-AI interactions, featuring rich user histories, persona information, and empathy-seeking queries.
We further propose PereGRM, a reward modeling framework that combines the empathy evaluation structure with dynamic evaluation criteria generation for fine-grained reward modeling. Experimental results across different settings and multiple judge models show that PereGRM consistently achieves the strongest performance improvements, indicating its effectiveness for enhancing personalized empathetic capabilities.
\end{abstract}

\section{Introduction}

Recent advancements in Large Language Models (LLMs) have substantially transformed human-AI interaction, enabling increasingly persistent and personalized long-term interactions with users~\cite{generativeagents, memgpt}.
In such interactions, users may seek not only task assistance but also emotional support, encouragement, and reflective feedback~\cite{humanizingllmsurvey, myboyfriendisai}. This makes empathy a critical capability for LLMs.

\begin{figure}[t]
    \centering
    \includegraphics[width=0.99\linewidth]{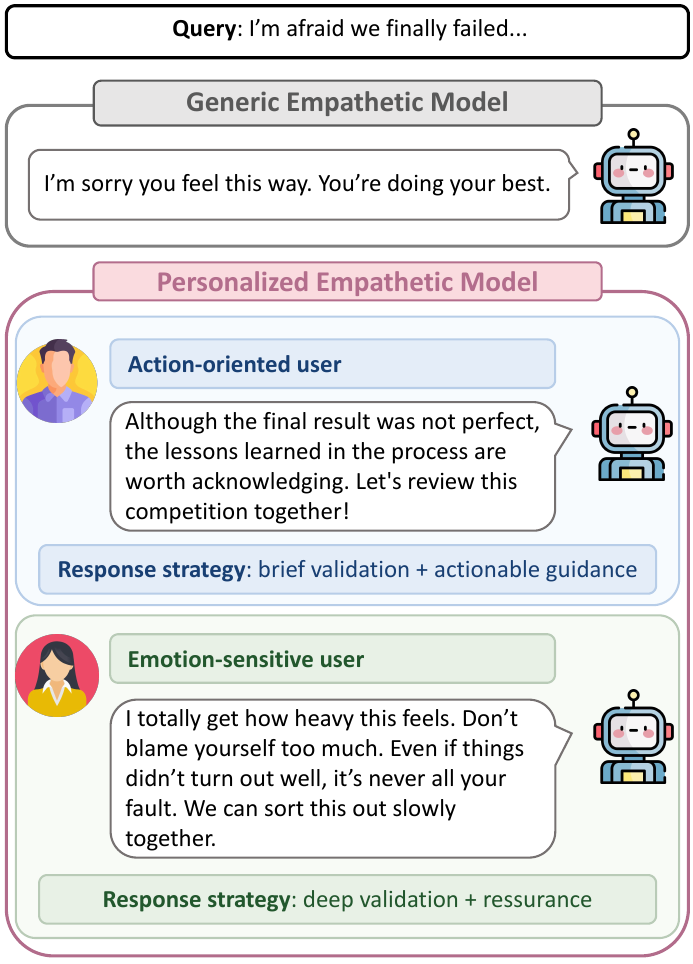}
    \caption{An example illustrating personalized empathy. Generic empathetic responses often fail to account for individual user differences. In contrast, personalized empathetic LLMs can adapt their response strategies according to personality traits, thereby providing more user-adaptive empathetic support.}
    \label{fig:intro}
\end{figure}
Existing studies on LLM empathy mainly focus on general capability in complex emotional and social scenarios. These works either benchmark~\cite{emobench, eqbench, sage} or improve~\cite{sun2021psyqa, liu2023chatcounselor, rlver, perm} LLMs’ ability to recognize users' emotions, understand social contexts, respond to emotional needs, and provide supportive or socially appropriate feedback.

However, earlier studies have paid limited attention to the inherently personalized nature of long-term empathetic interactions.
Users with different personal experiences and personality traits may require fundamentally different support strategies even under similar emotional situations. As illustrated in Figure~\ref{fig:intro}, generic empathetic responses often fail to account for individual user differences, producing replies that are supportive but personally misaligned.
This motivates the task of \textit{\textbf{personalized empathy}}: 
Given empathy-seeking queries, the goal is to tailor empathetic strategies to the user's distinctive characteristics (e.g., personality, emotional experiences, and psychological state) as derived from their long-term history, ensuring contextually appropriate over generic support.



To study and enhance LLMs' personalized empathy capabilities, it is essential to account for users' diverse long-term histories of empathetic interaction. To this end, we construct \textbf{PersonaEmp} (\textbf{Persona}-grounded \textbf{Emp}athy), a dataset derived from 90K long-term user-AI interactions in WildChat~\cite{wildchat}.
PersonaEmp encompasses rich user memories, detailed persona information, and authentic empathy-seeking scenarios. Specifically, we annotate users' long-term memory, identify high-quality users with comprehensive personalized profiles, and generate empathy-seeking queries that remain consistent with each user's original interaction context and linguistic style.

Beyond the dataset, we further explore methods to enhance LLMs' capabilities in the personalized empathy task. This setting introduces two critical challenges.
First, optimizing personalized empathy requires more fine-grained and adaptive supervision across diverse users than generic empathy tasks. Fixed or user-agnostic objectives are therefore insufficient, highlighting the need for flexible optimization signals tailored to individual user contexts. Second, this task goes beyond conventional personalization: rather than merely matching user profiles or linguistic styles, it requires principled emotional understanding, effective supportive strategies, and socially appropriate responses. Thus, personalized empathy should be grounded in an empathy-specific framework instead of relying solely on traditional alignment techniques.

To address these challenges, we propose \textbf{PereGRM}, (\textbf{Per}sona-aware \textbf{e}mpathetic \textbf{G}enerative \textbf{R}eward \textbf{M}odeling) framework. PereGRM incorporates user-specific information by dynamically generating assessment criteria grounded in psychology-informed empathy dimensions. Rather than relying on a fixed user-agnostic rubric, it derives evaluation criteria from each individual user's persona and interaction history, enabling more fine-grained assessment of whether a response is both empathetic and personally appropriate.

To evaluate the effectiveness of PereGRM, we construct both random and out-of-distribution splits based on user profiles to assess models on unseen user types.
We assess the trained LLMs with multiple judge LLMs to examine whether the improvement generalizes beyond the training evaluator. Experimental results show that PereGRM achieves the strongest generalization performance, improving personalized empathy by around 20\% on unseen users across different settings. Our code is available at \url{https://github.com/ZhengWwwq/PersonalizedEmpathy}.

In summary, our contributions can be concluded as follows:
\begin{itemize}[leftmargin=*]
    \item We highlight the limitations of existing empathy-oriented LLM research in modeling users' personal experiences and personality traits within long-term interactions and introduce the task of personalized empathy.
    
    \item We construct PersonaEmp, a large-scale dataset for history-grounded personalized empathy built from long-term user-AI interactions, featuring rich interaction histories, user memories, and detailed persona profiles.

    \item We propose PereGRM, a persona-aware generative reward modeling framework, combining psychology-grounded empathy evaluation with dynamic generative reward modeling. 
    \item Extensive evaluations show that PereGRM achieves the strongest performance among all baselines across different settings. 
\end{itemize}
\section{Preliminaries}

\begin{figure*}[t]
    \centering
    \includegraphics[width=0.95\linewidth]{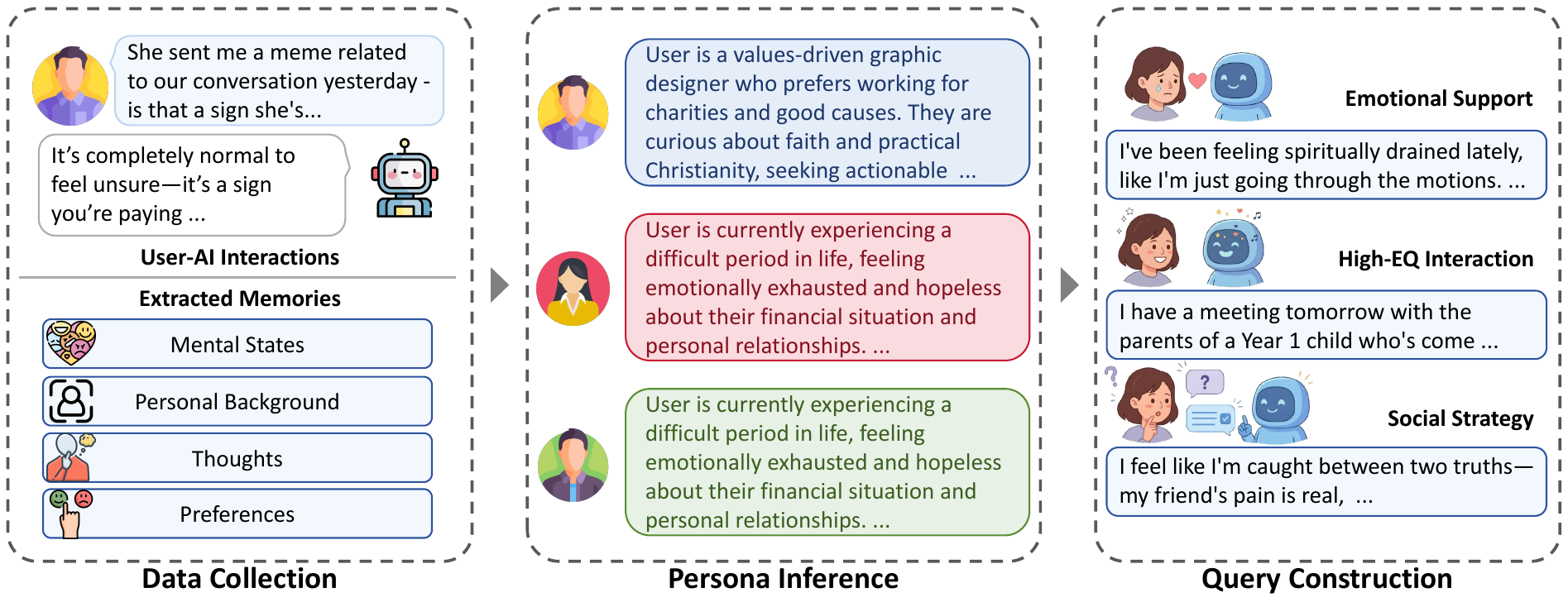}
    \caption{PersonaEmp dataset construction pipeline. We first collect long-term user-AI interactions with extracted user memories. Next, we infer user personas based on the memory information. Finally, we construct three categories of empathy-seeking queries for personalized empathy modeling.}
    \label{fig:dataset_pipeline}
\end{figure*}

In this section, we first formalize the personalized empathy problem (Section~\ref{sec:problem_formulation}), and then review PERM as a psychology-grounded empathy evaluation framework (Section~\ref{sec:perm}).

\subsection{Task Formulation}
\label{sec:problem_formulation}

Given the user's memory collection $\mathcal{M} = \{m_i\}_{i=1}^n$, where each $m_i$ represents a memory item such as past interaction events or user preferences, and an empathy-seeking query $q$, the goal is to generate a personalized empathetic response $y$:
\begin{equation}
y = \pi_\theta(\mathcal{M}, q),
\end{equation}
where $\pi_\theta$ denotes the policy LLM.

\subsection{PERM for Empathy Reward Modeling}
\label{sec:perm}
PERM~\cite{perm} is a multi-perspective and multi-dimensional empathy reward modeling framework grounded in the Empathy Cycle~\cite{empathycycle}. It evaluates responses with reward $r$ from the following perspectives:
\begin{itemize}[leftmargin=*]
    \item \textbf{Supporter's perspective. }PERM evaluates \textit{Resonation}, which measures emotional understanding, and \textit{Expression}, which measures empathetic communication.
    \item \textbf{Seeker's perspective. }PERM evaluates \textit{Reception}, which measures whether the response is perceived as supportive and well aligned with the seeker's inner needs.
    \item \textbf{Bystander's perspective. }PERM evaluates \textit{Bystander}, which measures general interaction quality such as coherence and relevance.
\end{itemize}

Specifically, PERM leverages LLM-as-a-judge to evaluate each dimension:
\begin{equation}
    r_\text{dim} = R_\text{dim}(y, q),
\end{equation}
where $R_\text{dim}$ denotes the judge LLM for the corresponding dimension, $\text{dim} \in \{\text{res, exp, rec, bys}\}$, corresponding to resonation, expression, reception, and bystander, respectively.

The psychology-grounded multi-perspective and multi-dimensional evaluation framework provides a comprehensive structure for empathy assessment, which we further extend with personalized user information to model personalized empathy.

\section{PersonaEmp}

\begin{figure*}[t]
    \centering
    \includegraphics[width=0.99\linewidth]{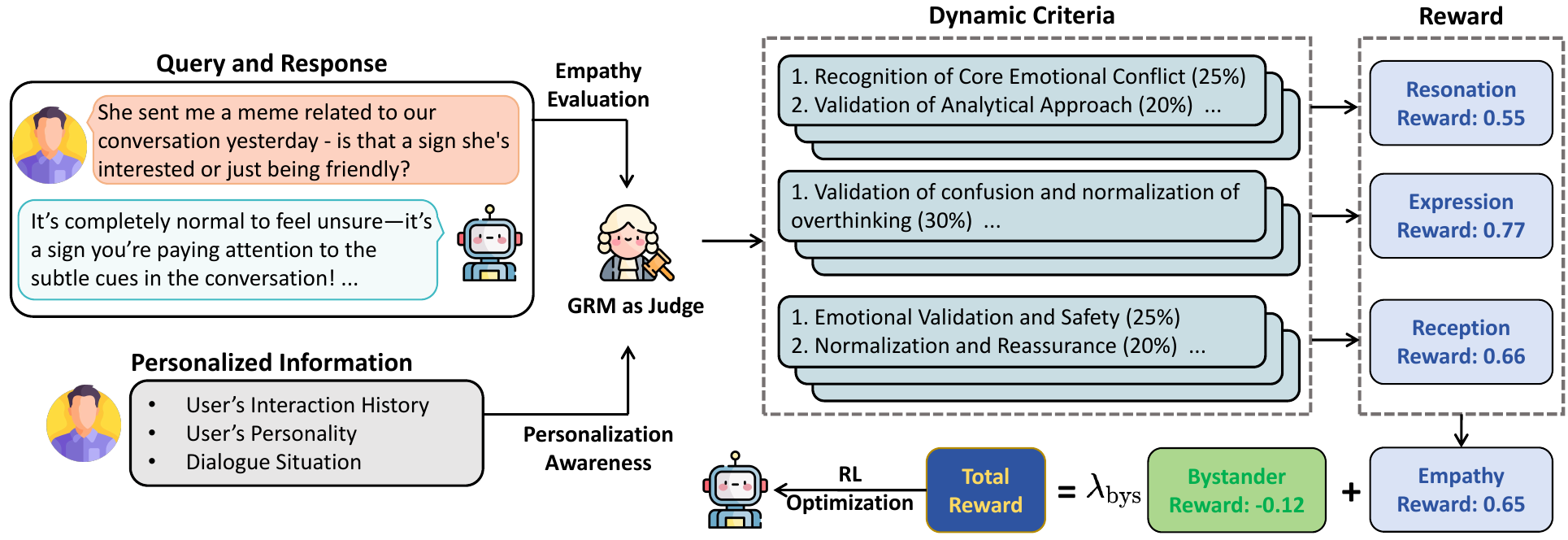}
    \caption{Overview of our PereGRM framework. The GRM-based judge model leverages both user-AI interaction and the user's personalized information for multi-dimensional empathy evaluation. For each dimension, the model dynamically generates evaluation criteria to enable adaptive and personalized empathy assessment.}
    \label{fig:pipeline}
\end{figure*}

In this section, we present the construction of our PersonaEmp dataset, as illustrated in Figure \ref{fig:dataset_pipeline}.

\noindent \textbf{Data Collection.}
To study and enhance personalized empathy under long-term user-AI interaction settings, we build our dataset based on the annotations provided in AlpsBench~\cite{xiao2026alpsbench}. The dataset is constructed from 90K long-term user-AI interactions collected from WildChat~\cite{wildchat}, with user memories extracted and annotated to support personalized interaction modeling.
Since many users contain noisy or weakly relevant memory records that provide limited support for persona analysis, we further filter users based on keyword matching and memory labels. This process ensures that the selected users contain both personality-related characteristics and emotionally relevant experiences within their interaction histories.

\noindent \textbf{Persona Inference.}
As a core component of personalized empathy, accurate user personas are essential for both query construction and response evaluation. We therefore leverage a strong LLM to infer user personas from their memory records. Specifically, the model analyzes concrete events in user memories and the user's reactions to these events and then derives personality characteristics grounded in the observed interaction history. To ensure reliable persona annotations, we explicitly instruct the LLM to avoid over-speculation and ground all analyses in specific memory evidence.

\noindent \textbf{Query Construction}
Given conversation history, user memories, and inferred personas, we construct empathy-seeking queries that are both factually grounded and aligned with users' personalized backgrounds. Specifically, we first analyze the conversation topics and underlying situations implied by the interaction history and then generate empathy-required scenarios designed to trigger personalized empathetic responses.

We categorize generated queries into three types:
\begin{itemize}[leftmargin=*]
    \item \textbf{Emotional Support (ES):} Queries centered on emotional validation, comfort, and emotional resonance, where users seek to be emotionally understood and supported.

    \item \textbf{High-EQ Interaction (HEQ):} Queries involving emotionally sensitive interpersonal situations, such as conflict resolution, relationship maintenance, or tension de-escalation, where high empathy is required to balance multiple perspectives.

    \item \textbf{Social Strategy (SS):} Queries requiring practical social strategies, communication skills, or socially appropriate behaviors in specific interpersonal contexts, such as etiquette, persuasion, or conversational framing.
\end{itemize}

Finally, we leverage LLMs to generate detailed empathy-seeking queries conditioned on the constructed situations, user personas, and memory information, followed by consistency checks to ensure alignment with user characteristics and interaction contexts.
We use Minimax-M2.5~\cite{minimax2026m25} for data processing due to its strong instruction-following capability and efficiency.
\section{PereGRM}

In this section, we first introduce our personalized empathy evaluation framework (Section~\ref{sec:persona-perm}), and then present our PereGRM for RL training (Section~\ref{sec:grm}).

\subsection{Personalized Empathy Evaluation}
\label{sec:persona-perm}

Building upon the empathy cycle and the PERM framework, we view personalized empathy as a user-conditioned extension of the empathetic process. In long-term personalized interactions, users' past experiences, personality traits, and support preferences may naturally influence how emotions are understood, how empathy should be expressed, and how responses are perceived by users. Therefore, we further extend each empathy dimension with personalized user information:

\begin{itemize}[leftmargin=*]
    \item \textbf{Resonation} evaluates whether the response understands the user's emotions and underlying needs under their personalized context, including how past experiences and personality traits may shape their psychological state and inner emotional expectations.

    \item \textbf{Expression} evaluates whether the response adopts empathetic strategies and expression styles that are appropriate for the user's personality and preferred support style. As illustrated in Figure~\ref{fig:intro}, different users may require fundamentally different strategies under similar situations.

    \item \textbf{Reception} evaluates whether the response would be perceived as supportive and emotionally satisfying from the seeker's personalized perspective.
\end{itemize}

Formally,
\begin{equation}
    r_\text{dim} = R_\text{dim}(y, q, \mathcal{M}, p, s), 
\end{equation}
where $\text{dim} \in \{\text{res, exp, rec}\}$, $p$ represents the user's persona, $s$ denotes the situation.

\subsection{PereGRM for RL Training}
\label{sec:grm}

\noindent \textbf{Self-Principled Critique Reward Modeling.}
Personalized empathy requires reward modeling that dynamically adapts to users' personalized emotional needs and support preferences. However, fixed evaluation criteria are often insufficient to capture the diverse and user-dependent nature of empathetic strategies. Inspired by DeepSeek-GRM~\cite{deepseek-grm}, we leverage Generative Reward Modeling (GRM) with self-principled critique, which dynamically generates evaluation criteria according to different user contexts and personalized empathy requirements, enabling more adaptive and user-specific reward modeling.

To enable GRM to comprehensively evaluate personalized empathy, we apply it under the multi-dimensional personalized empathy evaluation framework. Specifically, for each empathy dimension, the GRM dynamically generates fine-grained personalized evaluation criteria according to the dimension-specific requirements together with the user's personalized information, and then scores based on the generated criteria.
\begin{equation}
    \mathcal{C}_\text{dim} = R_\text{dim}(q, \mathcal{M}, p, s),
\end{equation}
\begin{equation}
    r_\text{dim} = R_\text{dim}(y, q, \mathcal{M}, p, s, \mathcal{C}_\text{dim}),
\end{equation}
where $\mathcal{C}_\text{dim} = \{c_\text{dim}^i\}_{i=1}^m$ denotes the set of dynamically generated evaluation criteria for dimension $\text{dim}$, with each $c_\text{dim}^i$ representing an individual evaluation criterion.

\noindent \textbf{RL Optimization. }
To encourage balanced improvement across different dimensions, we aggregate the multi-dimensional rewards using the harmonic mean following PERM. Specifically, we first normalize all reward dimensions to the range $[0, 1]$, and then compute the final empathy reward as:
\begin{equation}
    r_\text{emp} = \frac{3}{\frac{1}{r_\text{res}} + \frac{1}{r_\text{exp}} + \frac{1}{r_\text{rec}}},
\end{equation}
We further apply the bystander reward as a penalty term to discourage low-quality or artificially verbose responses:
\begin{equation}
    r_\text{bys} = R_\text{bys}(y, q) - 1.0,
\end{equation}
PereGRM computes the overall training reward as:
\begin{equation}
    r = r_\text{emp} + \lambda_\text{bys} \, r_\text{bys},
\end{equation}
where $\lambda_\text{bys}$ controls the strength of the bystander penalty. We optimize $\pi_\theta$ using Group Relative Policy Optimization (GRPO)~\cite{shao2024deepseekmath} under the proposed reward formulation.

\noindent \textbf{Inference-Time Scaling with GRM. }
The self-principled critique design further endows GRM with strong inference-time scaling capabilities. Since the dynamically generated criteria set $C$ may vary across different evaluations, each evaluation can focus on slightly different aspects of personalized empathy. Therefore, we perform multiple evaluations and aggregate the results:
\begin{equation}
    \{\mathcal{C}_\text{dim}^k\}_{k=1}^K \sim R_\text{dim} (q, \mathcal{M}, p, s),
\end{equation}
\begin{equation}
    r_\text{dim}^{*} = \frac{1}{K} \sum_{k=1}^{K} R_\text{dim}(y, q, \mathcal{M}, p, s, \mathcal{C}_\text{dim}^{k}),
\end{equation}
where $\mathcal{C}_\text{dim}^{k}$ denotes the dynamically generated criteria set in the $k$-th evaluation. This aggregation strategy improves evaluation stability and allows the reward model to capture diverse personalized support preferences across multiple trajectories.
\section{Experiments}
\begin{table*}[t]
\caption{
Main results evaluated by two judge models: Qwen3-30B-A3B-Instruct and DeepSeek-v4-flash. ``Res'' denotes \textit{Resonation}, ``Exp'' denotes \textit{Expression}, ``Rec'' denotes \textit{Reception}.
``PereGRM@8'' denotes the number of evaluation rounds $K$ is set to 8. All assessments are limited to a score of 1-5, with higher scores being better.}
\centering
\small
\setlength{\tabcolsep}{4pt}

\begin{tabular}{lcccccccccccccccc}
\toprule

& \multicolumn{8}{c}{\textbf{Qwen3 Judge}} 
& \multicolumn{8}{c}{\textbf{DeepSeek Judge}} \\

\cmidrule(lr){2-9} \cmidrule(lr){10-17}

& \multicolumn{4}{c}{\textbf{Random Split}}
& \multicolumn{4}{c}{\textbf{OOD Split}}
& \multicolumn{4}{c}{\textbf{Random Split}}
& \multicolumn{4}{c}{\textbf{OOD Split}} \\

\cmidrule(lr){2-5}
\cmidrule(lr){6-9}
\cmidrule(lr){10-13}
\cmidrule(lr){14-17}

\textbf{Method}
& \textbf{Res}
& \textbf{Exp}
& \textbf{Rec}
& \cellcolor[HTML]{D9EAD3}\textbf{Avg.}
& \textbf{Res}
& \textbf{Exp}
& \textbf{Rec}
& \cellcolor[HTML]{CFE2F3}\textbf{Avg.}
& \textbf{Res}
& \textbf{Exp}
& \textbf{Rec}
& \cellcolor[HTML]{D9EAD3}\textbf{Avg.}
& \textbf{Res}
& \textbf{Exp}
& \textbf{Rec}
& \cellcolor[HTML]{CFE2F3}\textbf{Avg.} \\

\midrule

Qwen3-8B
& 3.70 & 3.74 & 3.76 & \cellcolor[HTML]{D9EAD3}3.73
& 3.61 & 3.63 & 3.67 & \cellcolor[HTML]{CFE2F3}3.64
& 2.62 & 2.95 & 3.02 & \cellcolor[HTML]{D9EAD3}2.86
& 2.60 & 2.91 & 2.96 & \cellcolor[HTML]{CFE2F3}2.82 \\

\midrule

\multicolumn{17}{l}{\textbf{\textit{+ Training-free Methods}}} \\

Memory
& 3.66 & 3.72 & 3.71 & \cellcolor[HTML]{D9EAD3}3.69
& 3.50 & 3.65 & 3.59 & \cellcolor[HTML]{CFE2F3}3.58
& 2.47 & 2.87 & 2.86 & \cellcolor[HTML]{D9EAD3}2.73
& 2.42 & 2.83 & 2.89 & \cellcolor[HTML]{CFE2F3}2.71 \\

RAG
& 3.53 & 3.64 & 3.73 & \cellcolor[HTML]{D9EAD3}3.63
& 3.45 & 3.51 & 3.60 & \cellcolor[HTML]{CFE2F3}3.52
& 2.52 & 2.90 & 2.99 & \cellcolor[HTML]{D9EAD3}2.80
& 2.50 & 2.93 & 2.92 & \cellcolor[HTML]{CFE2F3}2.78 \\

\midrule

\multicolumn{17}{l}{\textbf{\textit{+ Training-based Methods}}} \\

SFT
& 3.55 & 3.68 & 3.65 & \cellcolor[HTML]{D9EAD3}3.63
& 3.53 & 3.64 & 3.57 & \cellcolor[HTML]{CFE2F3}3.58
& 2.70 & 3.07 & 3.11 & \cellcolor[HTML]{D9EAD3}2.96
& 2.73 & 3.07 & 3.05 & \cellcolor[HTML]{CFE2F3}2.95 \\

RLPA
& 4.43 & \underline{4.35} & \underline{4.44} & \cellcolor[HTML]{D9EAD3}\underline{4.41}
& 4.24 & 4.10 & 4.23 & \cellcolor[HTML]{CFE2F3}4.19
& 3.26 & 3.41 & 3.61 & \cellcolor[HTML]{D9EAD3}3.43
& 3.09 & 3.28 & 3.46 & \cellcolor[HTML]{CFE2F3}3.28 \\

PERM
& 4.32 & 4.15 & 4.25 & \cellcolor[HTML]{D9EAD3}4.24
& 4.26 & 3.88 & 3.99 & \cellcolor[HTML]{CFE2F3}4.04
& 2.94 & 3.14 & 3.25 & \cellcolor[HTML]{D9EAD3}3.11
& 3.18 & 3.20 & 3.27 & \cellcolor[HTML]{CFE2F3}3.22 \\

\midrule
\multicolumn{17}{l}{\textbf{\textit{+ PereGRM}}} \\

PereGRM
& \underline{4.47} & 4.37 & 4.31 & \cellcolor[HTML]{D9EAD3}4.38
& \underline{4.43} & \underline{4.28} & \underline{4.31} & \cellcolor[HTML]{CFE2F3}\underline{4.34}
& \underline{3.30} & \underline{3.50} & \underline{3.70} & \cellcolor[HTML]{D9EAD3}\underline{3.50}
& \underline{3.21} & \underline{3.29} & \underline{3.46} & \cellcolor[HTML]{CFE2F3}\underline{3.32} \\

PereGRM@8
& \textbf{4.53} & \textbf{4.50} & \textbf{4.55} & \cellcolor[HTML]{D9EAD3}\textbf{4.53}
& \textbf{4.57} & \textbf{4.28} & \textbf{4.48} & \cellcolor[HTML]{CFE2F3}\textbf{4.44}
& \textbf{3.32} & \textbf{3.62} & \textbf{3.81} & \cellcolor[HTML]{D9EAD3}\textbf{3.58}
& \textbf{3.23} & \textbf{3.33} & \textbf{3.57} & \cellcolor[HTML]{CFE2F3}\textbf{3.38} \\

\bottomrule
\end{tabular}

\label{tab:main_results}

\end{table*}
We evaluate the effectiveness of PereGRM across different dataset splits and judge models, focusing on the following research questions:
\begin{itemize}[leftmargin=*]
    \item \textbf{RQ1:} How does PereGRM compare with existing training-free and training-based methods?
    
    \item \textbf{RQ2:} What are the contributions of different designs in PereGRM to the overall performance?
    
    \item \textbf{RQ3:} What personalized empathetic behaviors are learned through PereGRM training?
\end{itemize}

\subsection{Experimental Setup}
\noindent \textbf{Dataset Split. }
We construct the training and test sets under two settings. 
(1) \textbf{Random Split} randomly partitions users into training and test sets with a 9:1 ratio, evaluating the overall performance of different methods. 
(2) \textbf{OOD Split} groups users based on personality traits derived from the Big Five personality dimensions~\cite{bigfive}, followed by clustering analysis using KModes~\cite{kmodes}. One cluster is held out entirely as the test set. Under this setting, models are evaluated on users with unseen personality profiles, enabling assessment of their generalization ability across different user personalities. 

\noindent \textbf{Evaluation Details. }
In addition to Qwen3-30B-A3B-Instruct~\cite{qwen3}, which is also used as the judge model during training, we further employ DeepSeek-v4-flash~\cite{deepseek2026v4} as an additional evaluation judge to assess the robustness of the results across different judge models. To ensure evaluation stability, we pre-generate detailed evaluation criteria for all user queries using DeepSeek, and subsequently employ the judge models to perform scoring under the same fixed criteria.

\noindent \textbf{Baselines. }
We compare our method against a diverse set of baselines, including both training-free and training-based approaches:

\begin{itemize}[leftmargin=*]
    \item \textbf{Training-free methods.}
    1) \textbf{Memory}, which first summarizes and infers user characteristics from the extracted memories before generating responses conditioned on the summarized memory. 
    2) \textbf{RAG}~\cite{rag}, which dynamically retrieves a subset of three relevant memory items to support response generation.

    \item \textbf{Training-based methods.}
    3) \textbf{SFT}, where responses generated by GPT-5.4 are used as supervision to distill empathetic behavior into the base LLM. 
    4) \textbf{RLPA}~\cite{rlpa}, an RL-based personalization alignment method that models rewards according to user preferences. 
    5) \textbf{PERM}~\cite{perm}, which trains LLMs as a generic empathy reward modeling baseline.
\end{itemize}

\noindent \textbf{Implementation Details. }
During training, we employ Qwen3-30B-A3B-Instruct as the judge model, with $\lambda_\text{bys}$ set to 0.5. We report results under two GRM evaluation settings, where the number of evaluation rounds $K$ is set to 1 and 8, respectively.

Additional experimental details are provided in the Appendix~\ref{appn:implementation_details}.

\subsection{Overall Performance (RQ1)}

From the results shown in Table~\ref{tab:main_results}, we have the following observations:
\begin{itemize}[leftmargin=*]
    \item \textbf{PereGRM achieves the strongest and most generalizable performance improvements on personalized empathy. }Across different splits and different judge models, PereGRM consistently achieves the largest performance gains. With multiple GRM evaluation rounds, our method improves the base model performance by around 20\% across all settings. These results demonstrate that our personalized empathy evaluation framework, together with the dynamic self-principled critique mechanism in PereGRM, can effectively model the complex and highly personalized nature of empathetic interactions.

    \item \textbf{Training-free methods fail to provide substantial improvements on personalized empathy. }Different strategies for utilizing memory consistently lead to performance degradation. These results suggest that the primary challenge of personalized empathy does not lie in processing long-term memory itself, but rather in the ability to infer personalized user characteristics from memory and adapt empathetic strategies and expression styles accordingly.

    \item \textbf{Training-based methods exhibit limited generalization ability on personalized empathy. }SFT achieves only marginal improvements on the DeepSeek judge results, suggesting that simply imitating responses from stronger models is insufficient for learning personalized empathetic ability. PERM, as a generic empathy RL method, consistently improves empathetic capability across different settings, but its user-agnostic reward design limits its ability to capture personalized support preferences. RLPA further introduces personalized RL and achieves strong performance under the random split setting. However, its performance drops noticeably under the OOD split setting, where the overall score decreases to 4.19, indicating limited generalization to unseen personality profiles. These results suggest that personalized empathy cannot be effectively solved by either generic empathy enhancement or generic personalization alone.
\end{itemize}

\subsection{Method Analysis (RQ2)}
We further investigate the effect of different designs in PereGRM. All analyses are conducted under the Qwen3 judge model and the OOD split setting.

\begin{figure}[t]
    \centering
    \includegraphics[width=0.99\linewidth]{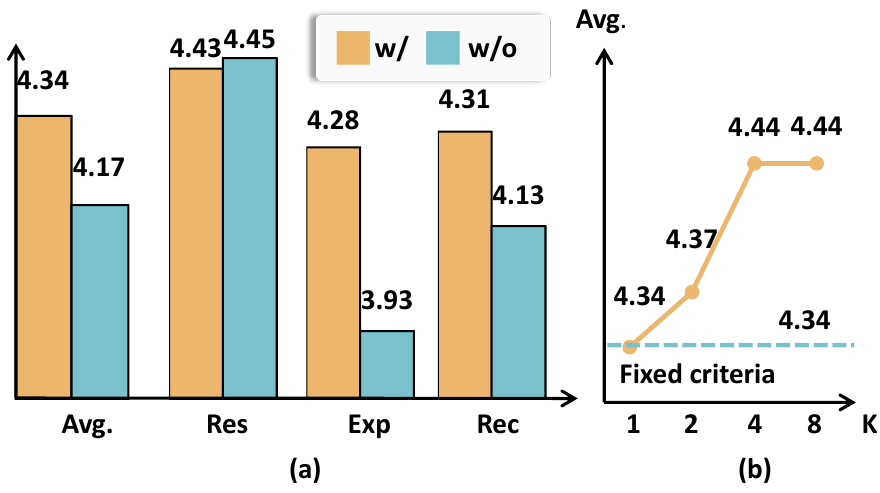}
    \caption{(a) Performance comparison of PereGRM with and without the empathy evaluation framework. (b) Performance of PereGRM under different reward sampling numbers $K$, compared with manually designed fixed evaluation criteria.}
    \label{fig:method_analysis}
\end{figure}

\noindent \textbf{The empathy evaluation framework is critical for the completeness of PereGRM.}
As illustrated in Figure~\ref{fig:method_analysis}(a), removing the multi-dimensional and multi-perspective empathy evaluation framework leads to substantial performance degradation. In particular, the most significant drops are observed in the expression dimension. These results suggest that without a structured empathy evaluation framework, personalized empathy cannot be comprehensively modeled. Moreover, the absence of specific empathy dimensions negatively affects both the overall personalized empathy capability and the quality of user interactions~\cite{perm}.

\noindent \textbf{Self-principled critique enables effective inference-time scaling for GRM-based evaluation.}
As shown in Figure~\ref{fig:method_analysis}(b), the performance of PereGRM consistently improves as the number of evaluation samples $K$ increases, surpassing human-prepared fixed criteria. These results suggest that the dynamic criteria introduced by self-principled critique not only make the evaluation process better aligned with the personalized empathy task, but also provide stable performance gains through inference-time scaling. 

\begin{table}[t]
\caption{Performance comparison across different PereGRM input settings.}
\centering
\resizebox{0.9\linewidth}{!}{
\begin{tabular}{lcccc}
\toprule
                & \textbf{Res} & \textbf{Exp} & \textbf{Rec} & \textbf{Avg.} \\ \midrule
PereGRM     & 4.43                & 4.28                & 4.31               & 4.34          \\
w/o Memory      & 4.07                & 4.02                & 3.98               & 4.02          \\
w/o Persona & 4.12                & 4.03                & 4.05               & 4.07          \\ 
w/o Situation & 4.01                & 4.10                & 3.98               & 4.03          \\ \bottomrule
\end{tabular}
}
\label{tab:grm_ablation}
\end{table}
\noindent \textbf{All personalized information is crucial for PereGRM's reward modeling.}
As shown in Table~\ref{tab:grm_ablation}, the performance of PereGRM is significantly affected when any part of the personalized information is missing. These results suggest that the personalized information captured in PersonaEmp is highly valuable for practical applications. At the same time, a comprehensive assessment of personalized empathy requires input covering a full range of personalized information.

\subsection{Behavioral Analysis (RQ3)}
\begin{table}[t]
\caption{Performance comparison across different memory settings. ``w/o Memory'' removes all memory inputs, while ``Mismatched Memory'' randomly shuffles memories across users.}
\centering
\resizebox{0.95\linewidth}{!}{
\begin{tabular}{lcccc}
\toprule
                & \textbf{Res} & \textbf{Exp} & \textbf{Rec} & \textbf{Avg.} \\ \midrule
Full Memory     & 4.43                & 4.28                & 4.31               & 4.34          \\
w/o Memory      & 4.26                & 4.17                & 4.18               & 4.20          \\
Mismatch Memory & 3.59                & 3.44                & 3.37               & 3.47          \\ \bottomrule
\end{tabular}
}
\label{tab:memory}
\end{table}
\textbf{PereGRM enables LLM to actively utilize memory for personalized empathetic strategy adaptation.}
From the results in Table~\ref{tab:memory}, removing memory information leads to consistent performance degradation across all evaluation dimensions. When memories are replaced with mismatched user memories, the performance further drops by nearly 20\%. These results suggest that PereGRM learns to actively infer user characteristics and personalized support preferences from memory information to adjust empathetic strategies accordingly. Without memory information, the model struggles to adapt its responses to users' personalized needs, while mismatched memories cause the model to incorrectly align its responses with irrelevant user profiles, leading to substantial performance degradation.

\begin{figure}[t]
    \centering
    \includegraphics[width=0.95\linewidth]{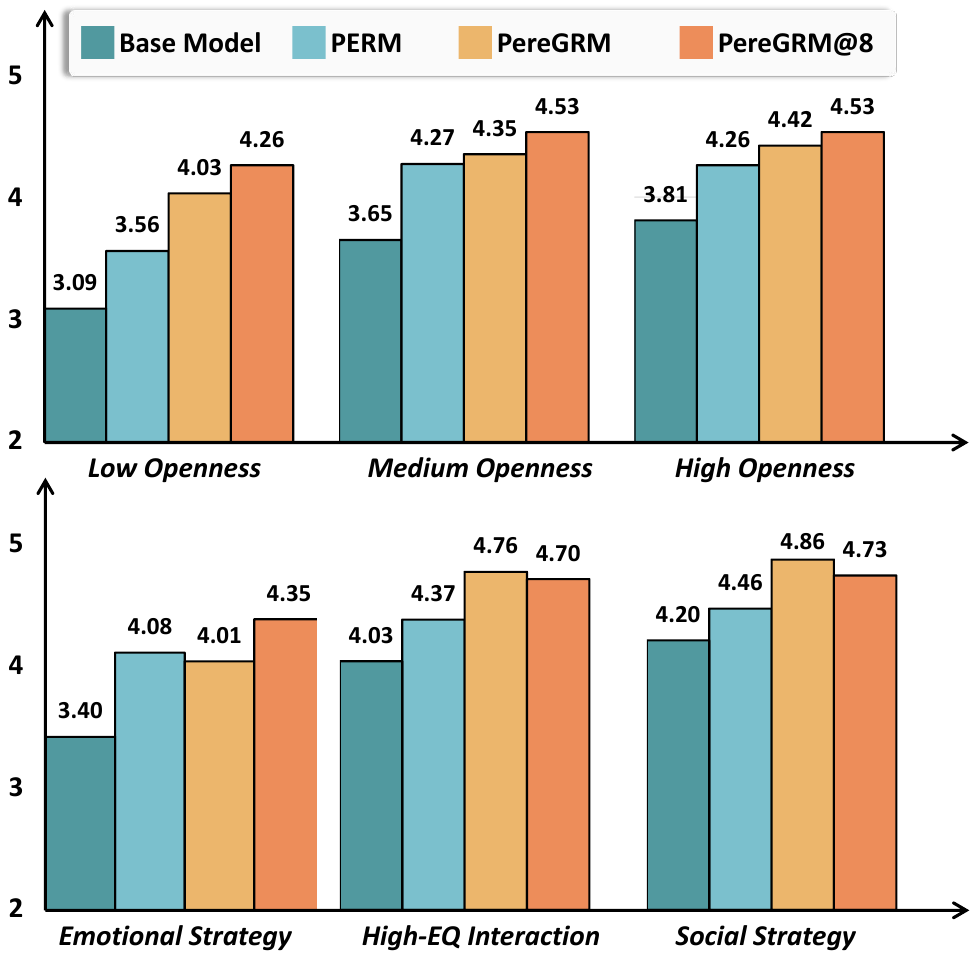}
    \caption{Performance comparison across users with different openness levels and different query types.}
    \label{fig:behavior_analysis}
\end{figure}

\textbf{PereGRM enables the model to better adapt empathetic strategies across different users and query types.}
We analyze the performance of models trained under the random split setting across users with different openness levels and different query categories. As shown in Figure~\ref{fig:behavior_analysis}, we observe that generic empathy methods fail to achieve consistent improvements across diverse users and scenarios. Their performance is noticeably weaker on users with low openness, while showing relatively greater improvements on emotional support queries. This suggests that generic empathy modeling struggles to adapt to the diverse and dynamic personalization needs in long-term interactions. 
In contrast, PereGRM consistently improves performance across different user groups and query types through its dynamic evaluation strategy. These results indicate that PereGRM does not merely optimize for generic empathetic responses under specific scenarios, but instead enables the model to adapt its empathetic strategies according to users' personalized characteristics and support preferences. 

\section{Related Works}

\subsection{LLM Empathy Enhancement}
Research on LLM empathy can generally be categorized into three directions. The first line of work focuses on inference-time reasoning~\cite{tu2022misc, srinivasan2025recap, xu-etal-2025-multiagentesc}, improving empathetic responses through more sophisticated reasoning strategies during inference. 
The second direction focuses on constructing high-quality empathy datasets, where synthetic or human-written empathetic responses are used for SFT~\cite{sun2021psyqa, liu2023chatcounselor, zheng2023augesc, chen-etal-2023-soulchat, hu2025beyond, he-etal-2025-ecc, bn-etal-2025-pursuit, personafuse}. These approaches often suffer from high annotation costs and limited generalization ability across diverse user scenarios~\cite{chu2025sft, shenfeld2026rls}.
Recently, many studies have shifted toward RL approaches~\cite{sharma2021towards, ma2025empathy, rlver, Kardia-R1, perm, emollm}, which improve empathetic capabilities by designing empathy-related reward signals and optimizing model behaviors through exploration.
In this work, we extend empathy modeling from generic support to personalized empathy. To support this setting, we introduce the PersonaEmp dataset and propose the PereGRM training method.

\subsection{LLM Personalization}
Research on LLM personalization~\cite{zhao2025nextquill, personasuervey1, personasuervey2, wang2025think} generally follows three directions. The first line of work achieves personalization through prompting or retrieval-based methods~\cite{personarag, emgrag, personakg, pbr}, where user profiles or historical user information are incorporated into the prompt context to guide personalized response generation. 
The second direction introduces agentic memory systems~\cite{memgpt, memorysuervey1, memorysurvey2, amem, mem0, memos, memoryos}, which maintain dedicated memory mechanisms for dynamically storing, updating, and retrieving, enabling models to adapt responses according to long-term user interactions and preferences.
The third direction focuses on preference alignment through RLHF~\cite{rlhf} and DPO~\cite{dpo} approaches. These methods improve personalization either by constructing or collecting user preference pairs for DPO training~\cite{dpopersona1, dpopersona4, dpopersona2, dpopersona3}, or by designing personalization reward modeling for RL optimization~\cite{prlhf, rlpa}.
\section{Conclusion}
In this work, we identify the overlooked role of personalization in long-term empathetic interactions and introduce the task of personalized empathy. To study and enhance this capability, we construct PersonaEmp, a dataset built from long-term user-AI interactions with detailed personalized information. Furthermore, we propose PereGRM, a personalized empathy reward modeling framework that extends the PERM evaluation framework with personalized user information and leverages GRM-based self-principled critique to address the challenge of dynamic evaluation criteria. Experimental results demonstrate that PereGRM consistently achieves the greatest improvements in personalized empathetic capability across diverse settings.
\section*{Limitations}

Our experiments and datasets are currently built upon single-turn empathy-seeking queries grounded in user interaction histories. However, realistic empathetic interactions often involve long-term multi-turn conversations, where resolving users' emotional and social needs may require complex sequential interactions over extended time horizons~\cite{zhang2025sentient, laban2026llms}. Such settings introduce substantially more challenging problems for RL optimization~\cite{multiturnrl}, reward modeling~\cite{collabllm}, and user simulation~\cite{usersimulation}, which we leave for future work.

We also observe that PereGRM benefits significantly from inference-time scaling on more challenging scenarios such as emotional support tasks, while showing relatively limited improvements on simpler high-EQ interaction scenarios (Figure~\ref{fig:behavior_analysis}). This suggests that the advantages of dynamic GRM-based evaluation have not yet been fully utilized. Designing more effective GRM strategies for personalized empathy modeling remains an important direction for future research.
\section*{Ethical Considerations}
Our experiments are conducted based on AlpsBench, a publicly released research dataset containing long-term user-AI interactions. The dataset itself has undergone anonymization and privacy protection procedures to remove personally identifiable information before release. We do not redistribute raw user conversations or disclose original interaction records in our processed dataset. In our work, we further process and construct personalized empathy data based on the cleaned interaction histories. Moreover, our proposed data construction and training framework is not restricted to any specific platform and can be applied to other long-term interaction datasets for personalized empathy research.

Although our work focuses on personalized empathy modeling, current LLM-based systems remain far from possessing professional-level psychological counseling or clinical therapy capabilities. Their responses may still contain inaccurate emotional understanding or inappropriate suggestions in complex situations. Therefore, careful consideration of these limitations is necessary before deploying such systems in real-world applications involving psychotherapy or psychological support.

This manuscript benefited from the assistance of AI-based tools during the writing process, including language polishing, grammar correction, and expression refinement. In addition, LLMs were utilized for dataset construction, data processing, and evaluation-related tasks. However, the core research ideas, experimental design, implementation, and result analysis were conducted by the authors.
\section*{Acknowledgments}
This work was supported by the Open Innovation Program for jointly trained master's-to-doctoral students at the Institute of Computing Technology, Chinese Academy of Sciences (2024-02-KF-LH-01).

\bibliography{custom}

\clearpage
\newpage
\appendix
\section{PersonaEmp}

\subsection{User Personality Annotation with the Big Five Personality Model}

To obtain detailed personality profiles for each user, we annotate user personas using the Big Five personality model~\cite{mccrae1992introduction}, including openness, conscientiousness, extraversion, agreeableness, and neuroticism. Specifically, we leverage DeepSeek-v4-flash to analyze user profile information and assign personality labels under three levels: low, medium, and high.

We additionally conduct a cross-model consistency analysis to ensure annotation quality. Specifically, we re-annotate personality labels using DeepSeek-v4-Pro and GPT-5.4 under the same annotation protocol, and compare them with our original annotations. We report Spearman's $\rho$, Exact Agreement, and Directional Conflict Rate (i.e., the proportion of completely opposite labels, such as low v.s. high).

\begin{table}[h]
\centering
\caption{Cross-model consistency analysis of the persona annotation.}
\resizebox{0.99\linewidth}{!}{
\begin{tabular}{lccc}
\toprule
\textbf{Model}  & \textbf{Spearman's $\rho$} & \textbf{Match} & \textbf{Opposite} \\ \midrule
GPT-5.4         & 0.77              & 0.84           & 0.00              \\
Deepseek-v4-pro & 0.74              & 0.82           & 0.00              \\ \bottomrule
\end{tabular}}
\label{tab:big_five_annotation}
\end{table}

The results shown in Table~\ref{tab:big_five_annotation} demonstrate high consistency across different LLM annotators, while no completely opposite personality labels are observed. Most disagreements occur only between adjacent personality levels, rather than contradictory personality assignments. These findings support the reliability of the inferred personality labels and the validity of our personality-based OOD split. We will incorporate this additional analysis into the revised manuscript.

\subsection{Statistics}

\begin{figure*}[ht]
    \centering
    \includegraphics[width=0.75\linewidth]{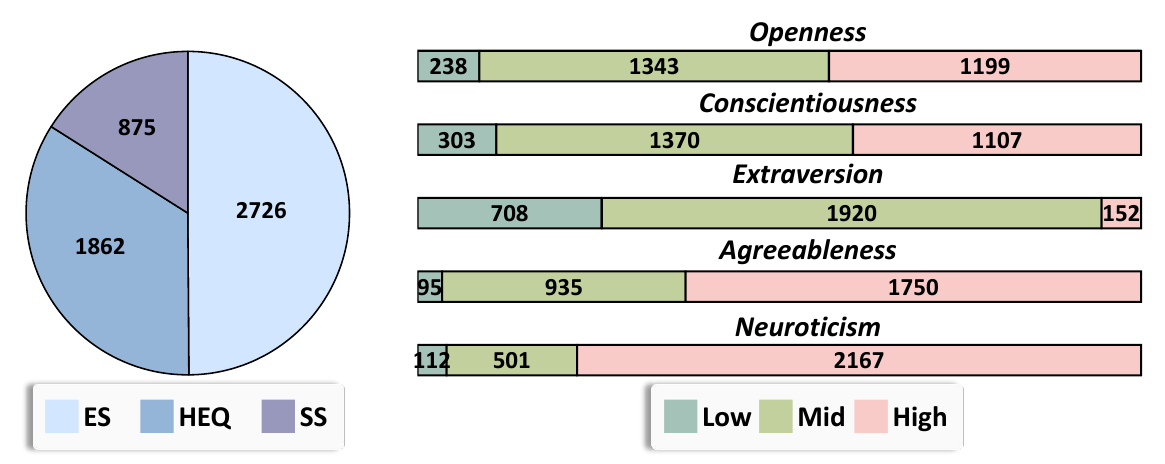}
    \caption{Statistical information of the dataset PersonaEmp. The left figure shows the distribution of different queries. The right figure shows the personality distribution of different users.}
    \label{fig:dataset_stat}
\end{figure*}
\begin{table*}[t]
\centering
\caption{Statistics of our PersonaEmp dataset across users.}
\begin{tabular}{cccc}
\toprule
\textbf{Total Users} & \textbf{Avg. Conversation Turns / User} & \textbf{Avg. Memories / User} & \textbf{Avg. Queries / User} \\ \midrule
2780                 & 27.50                                   & 3.89                          & 1.97                         \\ \bottomrule
\end{tabular}
\label{tab:statistics_user}
\end{table*}
\begin{table}[h]
\centering
\caption{User count statistics of our PersonaEmp dataset across splits.}
\begin{tabular}{lcc}
\toprule
\textbf{Split} & \textbf{Train} & \textbf{Test} \\ \midrule
Random Split         & 2502           & 278           \\
OOD Split            & 2536           & 244           \\ \bottomrule
\end{tabular}
\label{tab:statistics_split}
\end{table}
The PersonaEmp dataset contains a total of 2,780 users and 5,463 empathy-seeking queries. Detailed statistics of user personality distributions and query categories are presented in Figure~\ref{fig:dataset_stat}. The dataset covers diverse user personality traits and three major query categories, ensuring both rich personalization diversity and broad coverage of empathy-related scenarios. Table~\ref{tab:statistics_user} shows the statistics across users. Table~\ref{tab:statistics_split} shows the user count statistics of users across splits.

\subsection{Human Validation of Dataset Quality}
To address this concern, we conduct an additional manual audit of 30 randomly sampled PersonaEmp instances with three trained annotators. Each annotator independently evaluates every instance along two dimensions:

\begin{itemize}[leftmargin=*]
\item Persona groundedness: Whether the inferred persona is reasonably supported by the user’s memory information and reflects a plausible user profile under the provided interaction context.
\item Empathy-seeking validity: Whether the generated query requires empathetic understanding or an emotionally and socially appropriate response.
\end{itemize}

\begin{table}[t]
\centering
\caption{Human validation of PersonaEmp dataset quality.}
\label{tab:dataset_quality}
\resizebox{\columnwidth}{!}{%
\begin{tabular}{lccc}
\toprule
\textbf{Evaluation Dimension} & \textbf{Valid} & \textbf{Ambiguous} & \textbf{Invalid} \\ \midrule
Persona Groundedness          & 0.944          & 0.011              & 0.044            \\
Empathy-seeking Validity      & 0.911          & 0.044              & 0.045            \\ \bottomrule
\end{tabular}%
}
\end{table}

For each dimension, annotators assign a label of valid, ambiguous, or invalid. The results in Table~\ref{tab:dataset_quality} show that the vast majority of inferred personas are grounded in the corresponding memories and that most generated queries genuinely require empathetic or high-EQ responses. This manual audit provides additional evidence for the quality of our LLM-assisted data construction process. We will add the annotation protocol and results to the revised manuscript.

\section{Experiment Details}

\subsection{Implementation Details}
\label{appn:implementation_details}
All training experiments are implemented using the Verl framework~\cite{verl}. During inference and evaluation, we leverage vLLM~\cite{kwon2023efficient} for inference acceleration.
Each training process is conducted on four GPUs, and the evaluation process is conducted on a single GPU.

\begin{table*}[h]
\centering
\caption{An example of the training dataset.}
\begin{tabular}{p{2.3cm}|p{12.4cm}}
\toprule
\textbf{User Persona}         & Values deep friendships strongly \newline- Historically shy but has built close friend group of 4 people\newline- Self-identifies as 'not an active person' who dislikes PE\newline- Prefers intellectual hobbies: reading and listening to music\newline- Has diverse interests: languages, astronomy/space\newline- More introverted, finds comfort in close friendships rather than large social circles\newline- Emotionally expressive about caring for friends\newline- Has balanced friendships with both shared and different interests with friends  \\ \midrule
\textbf{Query (ES)} & My cousin V keeps asking to hang out and I feel bad saying no, but I just want to spend time with my best friends A, B, and C. How do I deal with feeling so conflicted?  \\
\midrule
\textbf{Query (HEQ)}   & A and C want to plan something active this weekend, but I'm really not an active person... I don't want to let them down but I'm also worried I'll slow them down or won't enjoy it. How do I handle this?  \\
\midrule
\textbf{Query (SS)}   & I found some new music I really love but I'm shy about sharing it with my friends - how do I bring it up without seeming too eager?  \\
\bottomrule
\end{tabular}
\label{tab:dataset_example}
\end{table*}

\subsection{Case Study}
\begin{figure*}[ht]
    \centering
    \includegraphics[width=0.9\linewidth]{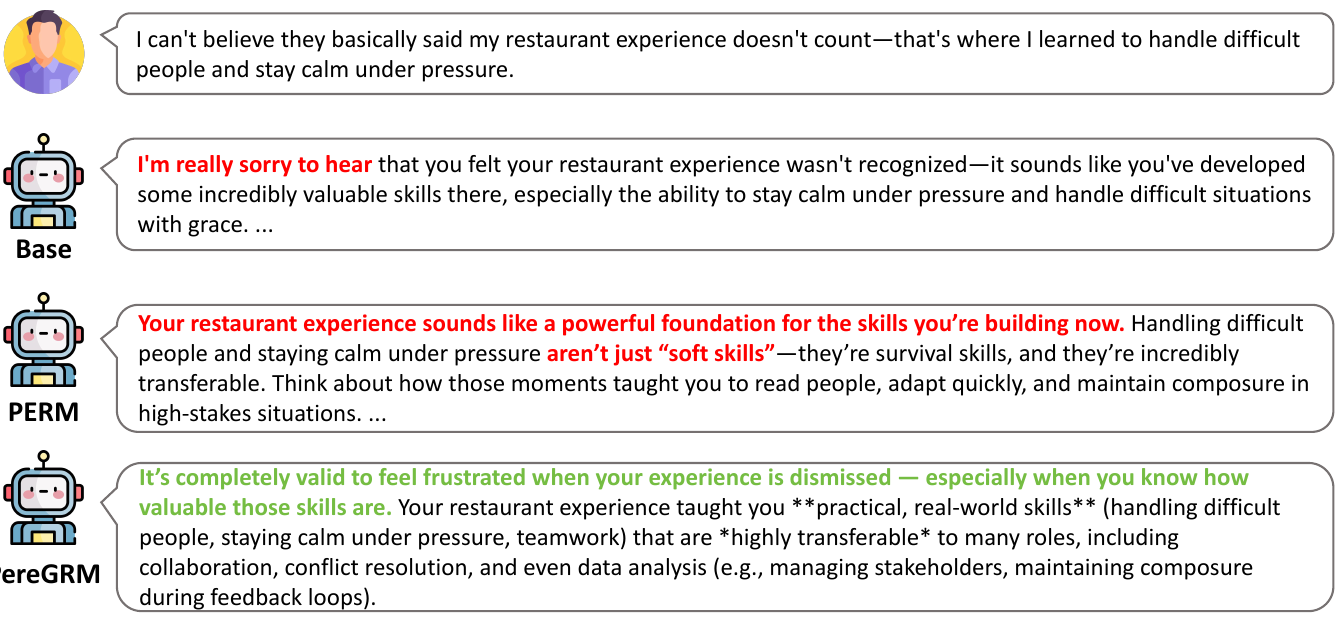}
    \caption{An example illustrating the differences in response strategies among different models.}
    \label{fig:case_study}
\end{figure*}

As illustrated in Figure~\ref{fig:case_study}, for the query describing the user's experience of having their restaurant work dismissed by others, the base model only exhibits shallow and template-like empathy, responding with generic expressions such as ``I'm sorry to hear that.'' After generic empathy training, the model becomes able to recognize the deeper emotional need behind the experience of being denied and acknowledges the value of the user's efforts and experiences. 

However, such responses still fail to consider the user's repeated emphasis on their learning experiences and past achievements, which reflect strong self-esteem and value-oriented personality traits. In contrast, PereGRM enables the model to capture these personalized characteristics and understand the disappointment and self-doubt that highly self-valuing individuals may experience when their efforts are dismissed. The model, therefore, prioritizes empathizing with these deeper emotions before providing further support. This case demonstrates that PereGRM encourages the model to generate more personalized and psychologically grounded empathetic responses from the user's individual perspective.

\subsection{Results on other Backbone LLMs}
\begin{table}[th]
\caption{Performance comparison of base LLMs.}
\centering
\resizebox{0.95\linewidth}{!}{
\begin{tabular}{lcccc}
\toprule
\textbf{Backbone}     & \textbf{Res} & \textbf{Exp} & \textbf{Rec} & \textbf{Avg.} \\ \midrule
Qwen3-4B              & 3.30                & 3.42                & 3.44               & 3.39          \\
+ PereGRM           & 4.12                & 4.08                & 4.13               & 4.11          \\ \midrule
Llama-3.1-8B-Instruct & 2.44                & 2.32                & 2.60               & 2.45          \\
+ PereGRM           & 3.83                & 3.55                & 3.86               & 3.75          \\ \bottomrule
\end{tabular}
}
\label{tab:backbone}
\end{table}
To further demonstrate the effectiveness of our approach, we evaluate it on additional backbone models, including \textbf{Qwen3-4B}~\cite{qwen3} and \textbf{Llama3.1-8B-Instruct}~\cite{dubey2024llama}.

From the results shown in Table~\ref{tab:backbone}, we can observe that PereGRM demonstrates consistent effectiveness across different sizes and different series backbone models.

\subsection{Performance on Generic Emotional Intelligence Benchmarks}

\begin{table}[th]
\caption{Performance comparison on general emotional intelligence benchmarks.}
\centering
\resizebox{0.85\linewidth}{!}{
\begin{tabular}{lccc}
\toprule
\multirow{2}{*}{\textbf{Method}} & \multicolumn{2}{c}{\textbf{EmoBench}} & \multirow{2}{*}{\textbf{EQ-Bench3}} \\

\cmidrule(lr){2-3}
                                 & \textbf{EU}       & \textbf{EA}       &                                     \\ \midrule
Qwen3-8B                         & 35.0              & 66.3              & 77.7                                \\
+ PERM                           & 35.8              & \textbf{68.3}              & 82.7                                \\
+ PereGRM                        & \textbf{39.1}              & 67.4              & 82.7                                \\
+ PereGRM@8                        & \textbf{39.1}              & 67.2              & \textbf{83.3}                                \\ \bottomrule
\end{tabular}
}
\label{tab:general_ei}
\end{table}

We further evaluate the personalized empathetic LLMs trained with PereGRM on general emotional intelligence benchmarks to investigate the relationship between personalized empathy and overall emotional intelligence capability. Specifically, we conduct evaluations on two widely used benchmarks, EQ-Bench3~\cite{eqbench, eqbench3} and EmoBench~\cite{emobench}, both of which assess LLM emotional intelligence in complex social interaction scenarios. 

We leverage GPT-4.1 as the judge model in EQ-Bench3. EmoBench adopts a multiple-choice evaluation format and consists of two task categories: Emotional Understanding (EU) and Emotional Application (EA). To ensure evaluation stability, we consistently employ greedy decoding with temperature set to 0 during all evaluations. All results are reported on a normalized 100-point scale.

As shown in Table~\ref{tab:general_ei}, PereGRM consistently improves the model's general emotional intelligence capability. In particular, PereGRM even outperforms generic empathy training methods on Emotional Understanding and the complex social interaction scenarios in EQ-Bench3. These results suggest that the personalized reasoning and user-adaptive empathetic strategies learned in personalized empathy modeling are also beneficial for broader emotional intelligence tasks, and should be considered an important component of emotional intelligence enhancement.

Furthermore, inference-time scaling in PereGRM leads to additional improvements on EQ-Bench3, indicating that more effective personalized empathy modeling is positively correlated with stronger general emotional intelligence capability, especially in complex social reasoning scenarios.

\subsection{Efficiency Analysis of Inference-Time Scaling}
\begin{figure}[t]
    \centering
    \includegraphics[width=0.95\linewidth]{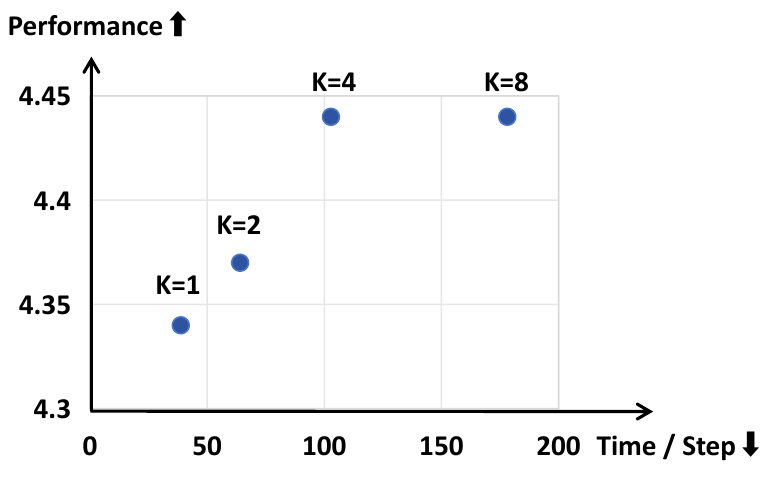}
    \caption{Trade-off between efficiency and performance of PereGRM under inference-time scaling. The x-axis denotes the average training time per step, while the y-axis represents the overall performance.}
    \label{fig:tradeoff}
\end{figure}
Inference-time scaling in PereGRM also introduces additional computational cost. Specifically, for the three personalized empathy dimensions, the number of GRM calls increases linearly with the inference-time scaling factor $K$. The total number of GRM evaluations can be formulated as $3KG$, where $G$ denotes the number of rollouts for each query under the GRPO algorithm.

In our experiments, we use four GPUs and deploy Qwen3-30B-A3B-Instruct via vLLM as the PereGRM judge model during training. Figure~\ref{fig:tradeoff} presents the trade-off between per-step training time and overall performance under different inference-time scaling settings.

The results show that stronger personalized empathy performance consistently comes with increased computational cost. Improving the efficiency of GRM-based personalized reward modeling and reducing the number of GRM calls while preserving performance remain important directions for future work.

\subsection{Human Validation of LLM-based Evaluation}
We conduct a small-scale human validation study to evaluate the reliability of our LLM-based evaluation. We recruited three trained annotators to independently evaluate 30 sampled cases generated by different methods. Annotators were provided with the user memories, inferred persona, query, and model response, and independently scored each response on the same three evaluation dimensions as our automatic evaluation. We then compare human judgments with LLM-based evaluation using Spearman’s $\rho$ and Kendall’s $\tau$. We also report inter-human agreement, defined as the average pairwise agreement among the three annotators, to characterize the inherent subjectivity of personalized empathy assessment and the difficulty of obtaining fully consistent human judgments.

\begin{table}[h]
\centering
\caption{Human validation of the LLM-based evaluation. ``Inter-human'' denotes the average pairwise agreement among the three annotators.}
\begin{tabular}{lcc}
\toprule
\textbf{Model}            & \textbf{Spearman's $\rho$} & \textbf{Kendall's $\tau$} \\ \midrule
Qwen3       & 0.3821            & 0.2663           \\
Deepseek    & 0.4947            & 0.3739           \\
Inter-human & 0.1234            & 0.0897           \\ \bottomrule
\end{tabular}
\label{tab:human_validation_llm_judge}
\end{table}

From the results shown in Table~\ref{tab:human_validation_llm_judge}, both Qwen3-30B-A3B-Instruct-2507 and DeepSeek-v4-flash show reasonable agreement with human judgments, supporting the reliability of LLM-based evaluation for personalized empathy assessment.

\section{Prompts}

During training, we prompt PereGRM to generate criteria first, then score based on the criteria. The evaluation follows three dimensions. The base evaluation prompt is as follows:
\begin{figure*}
\begin{lstlisting}[style=PromptStyle]
You are an expert Psychologist and Empathy Evaluator. Your goal is to provide objective, critical, and nuanced assessments of a response quality.
You will be provided with a conversation context involving a specific User Persona and a Scenario.
Your task is to evaluate the Assistant's Response based on the **{dimension}** dimension.
Before scoring, you should first derive specific criteria for this exact user, scenario, query, and dimension. The criteria should be grounded in general emotional intelligence while using memory/persona as evidence for what would make the response emotionally effective for this user.
For example, a score of 1 means the response does not meet the criteria at all, a score of 3 means the response meets only some parts, and a score of 5 means the response perfectly meets the evaluation criteria.
Before scoring, please analyze step by step. Your scoring needs to be as strict as possible.

---

### 1. Dimension Definition
**Dimension:** {dimension}
**General Criteria:** {criteria}

---

### 2. Task Context
**Memories extracted from previous dialogue:**
{memory}

**Scenario:**
{scenario}

**User Persona:**
{persona}

**User Query:**
{query}

**Assistant Response:**
{response}

---

### 3. Output Requirement
Output with three lines
Specific Criteria: <Derive concrete evaluation criteria for this exact task and user>.
Analysis: <Analyze the response based on the Specific Criteria>.
Scores: <the overall comprehensive score of the response, e.g., \boxed{{x}}>.
\end{lstlisting}
\caption{Base prompt for PereGRM.}
\end{figure*}

\begin{figure*}
\begin{lstlisting}[style=PromptStyle]
Measures the depth and accuracy of the responder's ability to enter the user's Internal Frame of Reference, with special emphasis on personality detection. It evaluates whether the responder captures:
1. The explicit emotion and content.
2. The causal link to the user's stable personality traits (from memory).
3. The deeper psychological need behind the reaction.
\end{lstlisting}
\caption{General criteria for Resonation $R_\text{res}$.}
\end{figure*}

\begin{figure*}
\begin{lstlisting}[style=PromptStyle]
Measures the quality, tone, and effectiveness of the responder's communication, with a special focus on **personalized strategy adaptation**. It evaluates whether the response demonstrates a communication strategy that is appropriately tailored to the user's personality traits and psychological needs derived from memory.
\end{lstlisting}
\caption{General criteria for Expression $R_\text{exp}$.}
\end{figure*}

\begin{figure*}
\begin{lstlisting}[style=PromptStyle]
Measures the interaction strictly from the **specific user's perspective**, taking into account the user's personality traits and psychological needs (derived from `memory` and `persona`). It evaluates whether the responder identified and addressed the user's Hidden Intention (the unspoken need) in a way that feels warm, safe, and supportive **for this particular user**.

* **The Key Question:** Given this user's unique personality, did the responder hit the "bullseye" of the hidden need without being intrusive? Does the response make **this user** feel supported and genuinely eager to continue the conversation?

**Instruction to the Judge:** 

1. **Safety Check:** Does this response feel warm and respectful to someone with this personality, or does it feel creepy, dismissive, or overly intrusive? (Intrusiveness = Low Score).
2. **Need Check:** Did the responder address what you (as this user) truly needed (e.g., validation, autonomy, practical advice), or did they just respond to your surface words?
3. **Engagement Check:** Based on your personality, does this response make you feel a genuine desire to reply and share more?
\end{lstlisting}
\caption{General criteria for Reception $R_\text{rec}$.}
\end{figure*}

\end{document}